# ESDiff: Encoding Strategy-inspired Diffusion Model with Few-shot Learning for Color Image Inpainting

Junyan Zhang, Yan Li, Mengxiao Geng, Liu Shi, Qiegen Liu, *Senior Member, IEEE*

*Abstract*—Image inpainting is a technique used to restore missing or damaged regions of an image. Traditional methods primarily utilize information from adjacent pixels for reconstructing missing areas, while they struggle to preserve complex details and structures. Simultaneously, models based on deep learning necessitate substantial amounts of training data. To address this challenge, an encoding strategy-inspired diffusion model with few-shot learning for color image inpainting is proposed in this paper. The main idea of this novel encoding strategy is the deployment of a "virtual mask" to construct high-dimensional objects through mutual perturbations between channels. This approach enables the diffusion model to capture diverse image representations and detailed features from limited training samples. Moreover, the encoding strategy leverages redundancy between channels, integrates with low-rank methods during iterative inpainting, and incorporates the diffusion model to achieve accurate information output. Experimental results indicate that our method exceeds current techniques in quantitative metrics, and the reconstructed images quality has been improved in aspects of texture and structural integrity, leading to more precise and coherent results.

## I. INTRODUCTION

Image inpainting is widely applied in fields of medical imaging, film restoration and creative editing [1-5]. It is aimed to provide visual and reasonable repair for images with missing, damaged, or contaminated areas [6-8]. It is challenging to generate the complete image from the exist image based on limited information.

In the past few decades, the technology has been fully developed because of the wide application of image inpainting [9, 10]. With the increasing interest of researchers in image inpainting technology, many research methods have been gradually proposed, which can be divided into two categories: traditional methods, and deep learning-based methods. Traditional methods usually include diffusion-based methods [11] [12] and patch-based methods [5, 13, 14]. Diffusion-based methods typically employ parametric models or partial differential equations to incorporate smooth priors. These approaches ensure that image contents smoothly diffuse from the boundaries of the missing region towards the interior, gradually synthesizing new textures and ultimately completing the inpainting process. Takeda *et al.* [15] proposed a method called kernel regression, in which the regression weights are calculated based on spatial proximity and intensity similarity. For patch-based methods, the best matching similar patches between the damaged area and other areas are found in the existing area according to the internal distribution of the existing image, and the patches are copied or spliced into the damaged area to repair the image. Aharon *et al.* [16] contributed the K-SVD algorithm to learn the sparse representation of each patch to capture local features. Jin and Ye [17] implemented image inpainting by constructing a Hankel matrix from image patches and maintaining the low-rank structure by annihilating filters (ALOHA). The patch-based methods are often limited by the number and quality of known areas, and are prone to mismatch with the search areas. Besides, it is difficult for this type of methods to obtain high-level semantic information of the image, resulting in semantic errors and edge breaks [18]. Classical diffusion-based methods, while effective for small gaps, tend to blur details in larger missing regions, lacking the sophistication of modern deep learning techniques that offer greater accuracy and visual coherence.

Deep learning has been widely applied in image processing in the recent years. To tackle the problem of global image distortion in patch-based methods, the decoder-encoder architecture of the CNN method was proposed [19]. Ulyanov et al. [20] demonstrated that randomly initialized networks can function as deep image priors (DIP), thereby bridging the gap between learning-based and non-learning-based recovery methods. The advent of diffusion-based generation models has further enhanced the effectiveness of image restoration. Examples include the denoising diffusion probability model (DDPM) [21], denoising diffusion implicit model (DDIM) [22], and latent diffusion model (LDM) [23]. GANs are increasingly favored by researchers due to their unique adversarial mechanisms, which enable models to better fit the data distribution of images[24]. Image restoration based on diffusion models generates images from noise and iteratively approximates the data distribution of images by progressively removing noise. These methods have been demonstrated to surpass the most advanced GAN-based methods in image synthesis [25]. Song *et al.* [26-28] developed a score-based generation model by integrating forward and backward diffusion processes into the stochastic differential equation (SDE) framework. This approach enables the generation of high-quality image samples without relying on adversarial optimization. To this end, they provided Noise Conditional Score Networks (NCSN) for learning and sampling from score-based models in high-dimensional space. Lugmayr *et al.* [29] developed an unconditional diffusion model that operates independently of mask shapes. In the reverse diffusion process, the model utilizes samples from known regions to iteratively reconstruct the image across multiple iterations, enhancing the

This work was supported in part by National Natural Science Foundation of China under 62122033 and Key Research and Development Program of Jiangxi Province under 20212BBE53001. (Junyan Zhang and Yan Li are co-first authors.) (Corresponding author: Liu Shi and Qiegen Liu.)

J. Zhang, Y. Li and M. Geng are with School of Mathematics and Computer Sciences, Nanchang University, Nanchang 330031, China ({zhangjunyan, liyanovo, mxiaogeng }@email.ncu.edu.cn).

L. Shi and Q. Liu are with School of Information Engineering, Nanchang University, Nanchang 330031, China ({shiliu, liuqiegen}@ncu.edu.cn).

coherence of the final output. Karras *et al.* [30] proposed to separate the diffusion scored-based models into multiple independent small components, which allows to modify a certain unit without affecting other components. In this framework, the author implemented an Ordinary Differential Equation (ODE) solver into the sampling process, which brought strong performance advantages. Significantly, unlike most diffusion models that require a large amount of training datasets to capture diverse features and scenarios, our method achieves effective modeling with significantly less data. This approach allows for robust performance even when data availability is constrained.

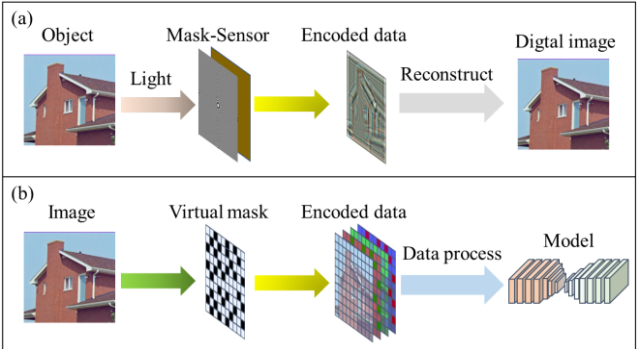

**Fig. 1.** The pipeline of coded mask imaging in (a) and our encoding strategy introduced by a "virtual mask" in (b).

Coded mask imaging involves incorporating a special designed mask within an imaging system, where incident light passes through aperture array of the mask. Multiple shifted and transformed encoded data are received by the sensor, which are reconstructed into the digital image using computational algorithms. The pipeline of coded mask imaging is illustrated in Fig. 1(a). Inspired by the concept of coded imaging, we propose a "virtual mask" that applies mutual perturbation to the RGB channels of the color image. By stacking the images obtained from these perturbations, high-dimensional encoded data are generated for image inpainting, as shown in Fig. 1(b).

In this paper, we propose a high-dimensional spatial diffusion model with few-shot learning for image inpainting by encoding strategy using a virtual mask (ESDiff). Specifically, our strategy involves implementing mutual perturbation transformations between high-dimensional channel information in color images using a virtual mask generated from a Gaussian distribution. This perturbation transformation leverages the virtual mask to account for color or spatial similarity of the same pixel in different channels in a color image, thereby obtaining more comprehensive prior information. For example, for each pixel, the change of its color value depends not only on the value of adjacent pixels, but also on the corresponding pixels of other channels. Therefore, through this recoding transformation, the model introduces diverse visual features and content representation methods, effectively leveraging the structure and content of the image to enhance the performance of image inpainting. The large amount of data results in higher computational costs, causing slower model training progress. The proposed few-shot learning strategy captures the features of the image, offering a solution to this challenge. We only need 10 image samples to train a sufficiently robust algorithm model. In summary, the main contributions of this work are as follows:

- ***The Construction of High-dimensional Objects via Virtual Mask.*** By leveraging the re-encoding of images introduced via virtual mask, our method can implement multiple mutual perturbation transformations between channels through the virtual mask to form a higher-dimensional object. This high-dimensional object can extract more effective information from the original data, thereby enhancing the robustness and accuracy of the derived insights.
- ***Few-shot Prior Learning from High-dimensional Objects.*** Enhanced by the construction of high-dimensional objects, our model can more efficiently extract features and facilitate representation learning. This innovative encoding strategy exploits the redundancy between channels to enable the network to understand complex image patterns more accurately and enhances generalization ability of the network with limited training samples.
- ***Multi-method Fusion in Iterative Inpainting.*** To make further use of the redundancy between channels, we incorporate the inverse mutual perturbation transformation into the diffusion model. Simultaneously, we introduce the low-rank method into the iterative reconstruction process, where its characteristic ability to identify consistent and predictable underlying image structures is utilized. Collectively, the integration of these techniques significantly enhances capability of the ESDiff model to restore both the intricate details and the overall structure of images.

The remainder of this paper is presented as follows. Section II briefly reviews some relevant works in this paper. In Section III, we elaborate on the formulation of the proposed method and describe the prior learning and iterative inpainting process of ESDiff in detail. Section IV evaluates the inpainting performance of the present method in comparison with other methods. Section V discusses some ideas about this work. Section VI will give the conclusion.

## II. RELATED WORK

### A. Image Inpainting

Generally, the formulation of image inpainting problem can be expressed as:

$$y = Dx + e \tag{1}$$

where $x \in \mathbb{R}^{M_x}$ is the unknown image subject to estimation., $y \in \mathbb{R}^{M_x}$ is the ruined image, $M_x$ denotes the dimensionality of the image, and $e$ represents additive noise. At the same time, $D \in \mathbb{R}^{M_x \times M_x}$ denotes the degradation matrix relating to a degraded imaging system. To elaborate, the operator $D$ is a binary mask that takes the value 0 if the pixel belongs to the missing or degraded information in the inpainting domain and $1$ otherwise. We assume $D^T D$ to be equal to $I$, where $I$ denotes an $M_x \times M_x$ identity matrix.

A classic approach determines the solution to Eq. (1) by solving the following constrained optimization problem:

$$\min_{x} \|Dx - y\|_2^2 + \lambda R(x) \tag{2}$$

where the first and second terms are the DC term and the regularization term, respectively. Then $\|\bullet\|_2$ denotes the $l_2$ norm and $\lambda$ is the balancing factor that adjusts the trade-off between the DC term and the regularization term.

To solve Eq. (2), various prior knowledge is incorporated into the regularization term to achieve stable and high-quality solutions. For instance, a sparsity-promoting regularizer derived from compressed sensing theory can be em-ployed [31], such as $l_0$ wavelet [32], $l_1$ wavelet [33], total variation [34], *etc.* Furthermore, the supervised end-to-end learning approaches typically adopt a discriminative framework to learn an implicit prior, which can be limited in terms of flexibility and robustness. In this study, we focus on constructing high-dimensional objects for prior learning using a score-based generative model.

*B. Score-based Generative Model with SDE*

Diffusion models, particularly those based on score-based Stochastic Differential Equations (SDE), have demonstrated significant success in generating realistic and diverse image tasks. The score-based SDE model includes two main processes: forward process and reverse process.

Specifically, a diffusion process $x(t) \in \mathbb{R}^{M_x}$ with $x(t) \in \mathbb{R}^{M_x}$ is indexed by a continuous-time variable $f \in \mathbb{R}^{M_x}$. There is a dataset of *i.i.d.* samples (*i.e.*, $x(0) \sim p_0$) and a tractable form to generate samples efficiently (*i.e.*, $x(T) \sim p_T$), where $p_0$ and $p_T$ refer to the data distribution and the prior distribution, respectively. Here, the diffusion process can be modeled as the equation of the following SDE:

$$dx = f(x,t)dt + g(t)dw \qquad (3)$$

where $f \in \mathbb{R}^{M_x}$ is the drift coefficient and $g \in \mathbb{R}$ is the diffusion coefficient of $x(t)$, while $w \in \mathbb{R}^{M_x}$ is the standard Wiener process.

By starting with samples of $x(T) \sim p_T$ and reversing the process, samples of $x(0) \sim p_0$ can be obtained. The above reverse process is also a diffusion process, which can be formulated as the reverse-time SDE:

$$dx = \left[ f(x,t) - g(t)^2 \nabla_x \log p_t(x) \right] dt + g(t) d\bar{w} \qquad (4)$$

where $dt$ is the infinitesimal negative time step, $\bar{w}$ is the reverse-time flow process of $w$, and $\nabla_x \log p_t(x)$ is the score of each marginal distribution. Since the true $s_\theta(x,t) \simeq \nabla_x \log p_t(x)$ of all $t$ is unknown, it can be estimated by training a time-dependent scoring network $s_\theta(x_t,t)$, *i.e.*, satisfy $s_\theta(x,t) \simeq \nabla_x \log p_t(x)$. Subsequently, samples can be generated by using a numerical SDE solver. Several general-purpose numerical methods exist for solving SDEs, such as the Euler-Maruyama method and stochastic Runge-Kutta method [35]. Any of these methods can be applied to the reverse-time SDE to facilitate sample generation.

### III. PROPOSED METHOD

In this section, we introduce the prior learning scheme and implementation details of the proposed ESDiff for color image inpainting.

*A. Motivation*

In recent years, deep learning has achieved significant advancements in the task of image inpainting [10, 36, 37], but it commonly necessitates a large amount of dataset. In many practical settings, particularly where data samples are limited, there is an increasing demand to train models to enhance their profound comprehension of the feature space from limited examples and to augment the generalization capability of the model from scant data volumes.

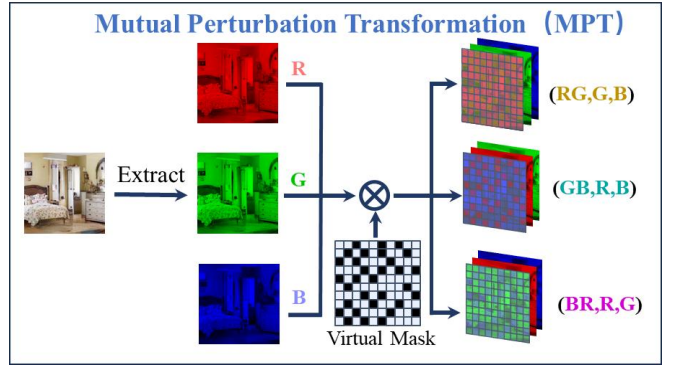

**Fig. 2.** The specific process of mutual perturbation transformation. We perturb a virtual mask generated by a normal distribution and use the redundancy between the channels of the image to bring more changes and diversity to the color coding of the image, generating a more complex image representation and training Produce higher-dimensional, higher-parameterized models.

A color images are typically represented in three distinct channels: the red, green, and blue (RGB) channels. There is a substantial correlation among three channels, as the intensity values across the red, green, and blue channels are interdependent and collectively determine the perceived color and luminance in an image [38]. We are committed to enhancing the inter-channel correlations to optimize feature learning in our model. Inspired by the concept of coded imaging, we propose a virtual mask generated based on a Gaussian distribution. Considering the inherent redundancy between color channels, we employ virtual masks to induce mutual perturbations among the channels. As shown in Fig. 2, our approach extends the original three-channel image into a high-dimensional tensor, thereby enhancing the feature representation of the image.

Block *et al.* [39] theoretically studied the $L_2$ bounds of the score estimation error of the generative modeling method. In their study, the representation boundary of the image is affected by the number of samples *n* and the feature dimension *d*. Specifically, suppose that $-\nabla \log_p$ is $M/2$-Lipschitz and $(m_{\sigma^2}, b_{\sigma^2})$-dissipative, then for $\sigma^2 \leq \sigma_{\max}^2 \leq m/2M$, there is a constant $B$ satisfies:

$$m_{\sigma^2} = \frac{m - \sigma^2 M}{2}, \qquad b_{\sigma^2} = b + \frac{B^2}{2(m - \sigma^2 M)} \qquad (5)$$

With probability at least $1 - 4\delta - Cne^{-R^2/m_t}$ on the randomness due to the sample, $\hat{s}(x)$ represents an empirical estimate of the score function $\nabla \log p_\sigma(x)$ of the smoothed distribution, so the expectation of loss can be expressed as:

$$E_{P_\sigma}\left[ \left\| \hat{s}(x) - \nabla \log p_\sigma(x) \right\|^2 \right] \\ \leq C(MR + B)^2 (\log^3 n \cdot \mathfrak{R}_n^2(\mathcal{F}) \cdot \beta_n d) \qquad (6)$$

where $C$ is a universal constant, $\mathfrak{R}_n^2(\mathcal{F})$ and $\beta_n$ contain the factor of $1/n$ and $\log\log(n)/n$, respectively.

From the formulation, as the number of samples *n* increases, the model achieves more consistent with the representation boundary of the dataset. Our work proposes an innovative method to make the encoding space more complex and diverse. By constructing high-dimensional data through virtual masks, the model can obtain more comprehensive image distribution information, thereby facilitating the accurate learning of image features and structures with limited training datasets. Precisely, the virtual mask can be regarded as a virtual matrix $M_{ij}$, so the corresponding channels can be mutually perturbed by the virtual mask $M_{ij}$. The mutual perturbation transformation is shown in the Fig. 2. Three different ways to perturbated can be described separately as:

$$MPT : \begin{cases} C_{RG} = C_R * M_{ij} + C_G * (1 - M_{ij}) \\ C_{GB} = C_G * M_{ij} + C_B * (1 - M_{ij}) \\ C_{BR} = C_B * M_{ij} + C_R * (1 - M_{ij}) \end{cases} \quad (7)$$

After the mutual perturbation transformation, we stack the original image, RG perturbation tensor, GB perturbation tensor, and BR perturbation tensor to form a high-dimensional tensor. In this work, the stacked tensor $X_c = \{[R,G,B],[RG,G,B],[R,GB,B],[R,G,BR]\}$ with three types of MPT operator.

While, the inverse-mutual perturbation transformation process is to restore the perturbed channel to the original channel, could be given as follow:

$$i\text{-}MPT : \begin{cases} C_R = C_G * M_{ij} + C_{RG} * (1 - M_{ij}) \\ C_G = C_B * M_{ij} + C_{GB} * (1 - M_{ij}) \\ C_B = C_R * M_{ij} + C_{BR} * (1 - M_{ij}) \end{cases} \quad (8)$$

Additionally, the score-based model is considered capable of capturing the internal statistical distribution of the data. Consequently, utilizing a score-based generative model to mathematically model the cut patches in the sample is seen as beneficial, as it facilitates learning the prior information pertinent to the internal structure of the image. In the iterative inpainting process, the mutual perturbation transformation across the channels of the virtual mask can be regarded as a re-encoding of the original image. By modeling the data distribution within the latent space through parameterized diffusion models, ADMM is employed to leverage its capability in preserving the sparsity of the underlying image structure. Furthermore, the proposed method integrates data consistency assurance to maintain high fidelity to the original data while employing more complex parameterized diffusion models, thereby acting as a constraint to guide the reconstruction process. In summary, conditional sampling is executed through a sequence of steps: forward perturbation, Predictor-Corrector (PC) sampler, reverse perturbation, LR step, and DC step. The detailed mechanisms of the prior learning and iterative inpainting strategy in ESDiff will be elucidated in the following two sections.

*B. ESDiff: Prior Learning*

In this work, image data is constructed as patches randomly cut from the original image for previous learning, rather than the most common method of training in the image domain.

***Extracting Internal-Middle Patches from Images.*** Rather than traditional methods which require a large number of training samples, the proposed method only requires a small number of images for training. To accommodate local image statistics, these original images are randomly cropped into many internal-middle patches $x_n$ with size of $64 \times 64 \times 3$. Note that at each iteration of the training process, an internal-middle patch needs to be extracted from each of these images.

***Mutual Perturbation Transformation (MPT) in Prior Learning.*** For the $64 \times 64 \times 3$ patches randomly extracted from the original image due to its natural redundancy between RGB channels, we innovatively adopt an encoding transformation method called channel perturbation transformation to re-encode the data. We introduce a virtual mask that matches the patch size and is randomly generated from a normal distribution. Using this virtual mask, we perturb the R-G, R-B, and G-B channels relative to each other to create patches with more complex coding patterns

***Generative Modeling on Perturbated Patches.*** Afterwards, the perturbated patches are used as the input of the score-based network for training. We use a diffusion model to perform prior learning on the distribution of unstructured data after transformation, in order to achieve a higher parameterized model.

In detail, the prior learning stage in the score-based model is manipulated by employing an SDE to transform a complex data distribution into a well-defined prior distribution. Empirically, variance exploding (VE) SDE [28] typically produces higher quality samples. It chooses $f(x,t) = 0$ and $g(t) = \sqrt{d[\sigma^2(t)]/dt}$ for Eq. (3). The forward VE-SDE can be expressed as:

$$d\mathcal{X} = \sqrt{\frac{d[\sigma^2(t)]}{dt}} dw \quad (9)$$

where $\sigma(t)$ is Gaussian noise function with a variable in continuous time $t \in [0,1]$, which can be reinterpreted as a positive noise scale $\{\sigma_i\}_{i=1}^N$. The forward VE-SDE can be represented as a Markov chain in the model training stage: $\mathcal{X}^{i+1} = \mathcal{X}^i + \sqrt{\sigma_{i+1}^2 - \sigma_i^2} z, \ i = 0, \cdots, N-1$.

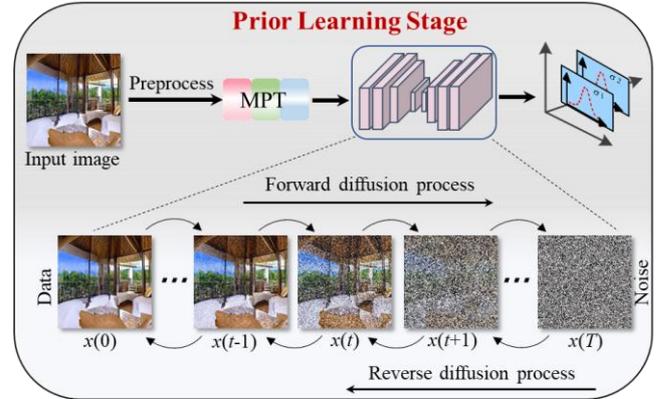

**Fig. 4.** The pipeline of the prior learning stage in ESDiff. Before the prior learning of the score-based network, internal-middle patches are extracted from the original images and then perform various perturbations on the original patches. After the training procedure is finished, the network $s_\theta(\mathcal{X}(t),t)$ learns the data score $\nabla_\mathcal{X} \log p_t(\mathcal{X})$.

Since $\nabla_\mathcal{X} \log p_t(\mathcal{X}(t))$ is intractable for all $t$, it can be estimated by training a score-based model using denoising score matching on samples. To achieve this, the time-dependent score-based model $s_\theta(\mathcal{X}(t),t)$ can be trained by optimizing the parameters $\theta$:

$$\theta^* = \arg\min_\theta \mathbb{E}_{t \sim U(0,1)} \Big\{ \gamma(t) \mathbb{E}_{\mathcal{X}(0)} \mathbb{E}_{\mathcal{X}(t)|\mathcal{X}(0)} \Big[ \\ \|s_\theta(\mathcal{X}(t),t) - \nabla_\mathcal{X} \log p_{0t}(\mathcal{X}(t)|\mathcal{X}(0))\|_2^2 \Big] \Big\} \quad (10)$$

where $\gamma : [0,T] \to \mathbb{R}_{>0}$ serves as a positive weighting function and $t$ is uniformly sampled from the interval $[0,T]$. $p_{0t}(\mathcal{X}(t)|\mathcal{X}(0))$ represents the Gaussian perturbation ker-nel centered at $\mathcal{X}(0)$. Once the network could achieve convergence to $s_\theta(\mathcal{X}(t),t) \simeq \nabla_\mathcal{X} \log p_t(\mathcal{X}(t))$, it indicates that $\nabla_\mathcal{X} \log p_t(\mathcal{X}(t))$ is known for all $t$ by $s_\theta(\mathcal{X}(t),t)$. Thus, it allows us to derive the reverse-time SDE and simulate it in sample, which can be given by:

$$d\mathcal{X} = -d\sigma^2(t) \bullet \nabla_\mathcal{X} \log p_t(\mathcal{X}) + \sqrt{d\left[\sigma^2(t)\right]/dt}\, d\bar{w} \quad (11)$$

Fig. 4 provides a comprehensive description of the bidirectional process of the SDE. Firstly, the samples are constructed into high-dimensional objects though mutual perturbation transformation operators. Secondly, the forward VE-SDE diffuses the image data distribution $\mathcal{X}(0) \sim p_0$ into a prior distribution $\mathcal{X}(T) \sim p_T$. Then, to estimate the score of the distribution at each intermediate time step, which can estimate a score function with a time-conditional neural network $s_\theta(\mathcal{X}(t),t) \simeq \nabla_\mathcal{X} \log p_t(\mathcal{X}(t))$. Finally, by reversing the forward process and starting from samples of $\mathcal{X}(T) \sim p_T$, samples $\mathcal{X}(0) \sim p_0$ can be obtained.

### C. ESDiff: Iterative Inpainting

This study deeply explores the encoding strategy, constructs a perturbation module and inverse-perturbation module and applies it in an iterative loop. In this work, the proposed ESDiff utilizes a score-based generative model and flexibly incorporates a low-rank step to improve the accuracy and robustness of inpainting. The specific implementation process of restoring images in $256 \times 256 \times 3$ can be divided into the following steps:

***Extracting Internal-middle Patches.*** In order to facilitate processing of the edge area of the image during patching processing, the image is padded to a size of $320 \times 320 \times 3$, and then patch in sequence to extract 25 patches of size $64 \times 64 \times 3$ (*i.e.*, $\{x_n, n = 1, \cdots, 25\}$) from the padded image.

***Mutual Perturbation Transformation (MPT) on Patches.*** In order to enhance the ability of model to capture the texture details and structure of the image, ESDiff introduces encoding transformation into the image inpainting process. The perturbation process in this module can be referred to Section III. B and can be expressed as $\mathcal{X}_{MP} = MPT(\mathcal{X}_n)$. The perturbation process is consistent with the two-by-two perturbation method between the three channels *R*, *G*, and *B* in the training process.

***Conditional Generation.*** At the conditional generation process, PC sampler, inverse-mutual perturbation transformation (i-MPT), LR step, and DC step are alternatively performed.

As introduced in the section III. B, the forward VE-SDE process allows us to produce data samples not only from $p_0$, but also from $p_0\left[(\mathcal{X}(0), \mathcal{X}_{MP})|Y\right]$ as $p_t\left[Y|(\mathcal{X}, \mathcal{X}_{MP})\right]$ is known. Thus, we can sample from $p_t\left[(\mathcal{X}, \mathcal{X}_{MP})|Y\right]$ by starting from $p_T\left[(\mathcal{X}(T), \mathcal{X}_{MP})|Y\right]$. In other words, the proposed ESDiff interprets the observation as a conditional generation within a low-rank framework and integrates it into the iterative inpainting process, i.e.,

$$\nabla_\mathcal{X} \log p_t\left[(\mathcal{X}, \mathcal{X}_{MP})|Y\right] \simeq \nabla_\mathcal{X} \big[\log p_t(\mathcal{X}|\mathcal{X}_{MP}) \\ + \log p_t(\mathcal{X}_{MP}) + \log p_t\left[Y|(\mathcal{X}, \mathcal{X}_{MP})\right]\big] \quad (12)$$

where $Y$ is the measurement corresponding to $\mathcal{X}$. By substituting Eq. (12) into the corresponding part of Eq. (11), the conditional sampling is performed by solving a conditional reverse-time VE-SDE. The discretized form of this VE-SDE can be represented as:

$$\mathcal{X}_n^i = \mathcal{X}_n^{i+1} + (\sigma_{i+1}^2 - \sigma_i^2)\nabla_\mathcal{X}\big[\log p_t(\mathcal{X}_n^{i+1}|\mathcal{X}_{MP}^{i+1}) + \log p_t(\mathcal{X}_{MP}^{i+1}) \\ + \log p_t\left[Y_n|(\mathcal{X}_n^{i+1}, \mathcal{X}_{MP}^{i+1})\right]\big] + \sqrt{\sigma_{i+1}^2 - \sigma_i^2}\, z \quad (13)$$

Eq. (13) can be decoupled into three sub-problems: The measurement distribution term with low-rank conditional $\log p_t(\mathcal{X}|\mathcal{X}_{MP})$, the prior information of samples perturbed by virtual mask $\log p_t(\mathcal{X}_{MP})$, and the DC step $\log p_t\left[Y|(\mathcal{X}, \mathcal{X}_{MP})\right]$. Consequently, the conditional generation can be performed by alternatively iterating the PC sampler, i-MPT step, LR step and DC step. The details are demonstrated as follow:

***PC Sampler.*** According to Section III. B, $s_\theta(\mathcal{X}(t),t)$ trained with denoising score matching almost satisfies $s_\theta(\mathcal{X}(t),t) \simeq \nabla_\mathcal{X} \log p_t(\mathcal{X}|\mathcal{X}_{MP})$. In this work, the PC sampler is introduced to correct errors in the evolution of the discretized reverse-time SDE.

Specifically, the predictor refers to a numerical solver for the reverse-time SDE, which generates an estimate of the sample at the subsequent step:

$$\mathcal{X}_n^i = \mathcal{X}_n^{i+1} + (\sigma_{i+1}^2 - \sigma_i^2)s_\theta(\mathcal{X}_n^{i+1},t) + \sqrt{\sigma_{i+1}^2 - \sigma_i^2}\, z \quad (14)$$

where $i = N-1, \cdots, 1, 0$ is the number of discretization steps for the reverse-time SDE, $\sigma_i$ is the noise schedule at the $i^{th}$ iteration, and $z \sim \mathbb{N}(0,1)$ is the standard normal.

Simultaneously, the corrector is defined as the iteration procedure of Langevin dynamics, employed to correct the marginal distribution of the estimated sample:

$$\mathcal{X}_n^{i,j} = \mathcal{X}_n^{i,j-1} + \varepsilon_i s_\theta(\mathcal{X}_n^{i,j-1},t) + \sqrt{2\varepsilon_i}\, z \quad (15)$$

where $j = 1, 2, \cdots, M$ is the number of corrector steps, $\varepsilon_i > 0$ is the noise step size at the $i^{th}$ iteration. During the above iterative procedure, when $N \to \infty$, $M \to \infty$ and $\varepsilon_i \to 0$, $\mathcal{X}_n^{i,j}$ is a sample from $p_t(\mathcal{X}_{MP})$ under specific predefined perturbation conditions.

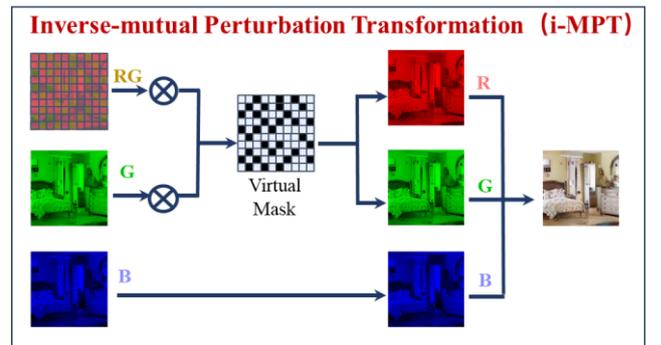

**Fig. 5.** Reverse restoration process for the *RG* perturbation case. Through the saved virtual mask and the *R* channel image and *G* channel image disturbed by the *G* channel, we can restore the original image information.

***Inverse-mutual Perturbation Transformation (i-MPT).*** To enhance the effectiveness of the encoding strategy during both training and testing phases, the model incorporates an inverse transform module following the PC sampler in the iterative reconstruction process. This module supports three types of perturbation transformations: *RG* perturbation, *GB* perturbation, and *BR* per-turbation. In the inverse transform module, the transformation can be accurately reversed using the corresponding transformation method and the saved mask. Specifically, for the *RG* perturbation method, the pixels of the *RG* channel and the *G* channel are swapped according to the mask. The process of this reverse perturbation is illustrated in Fig. 5.

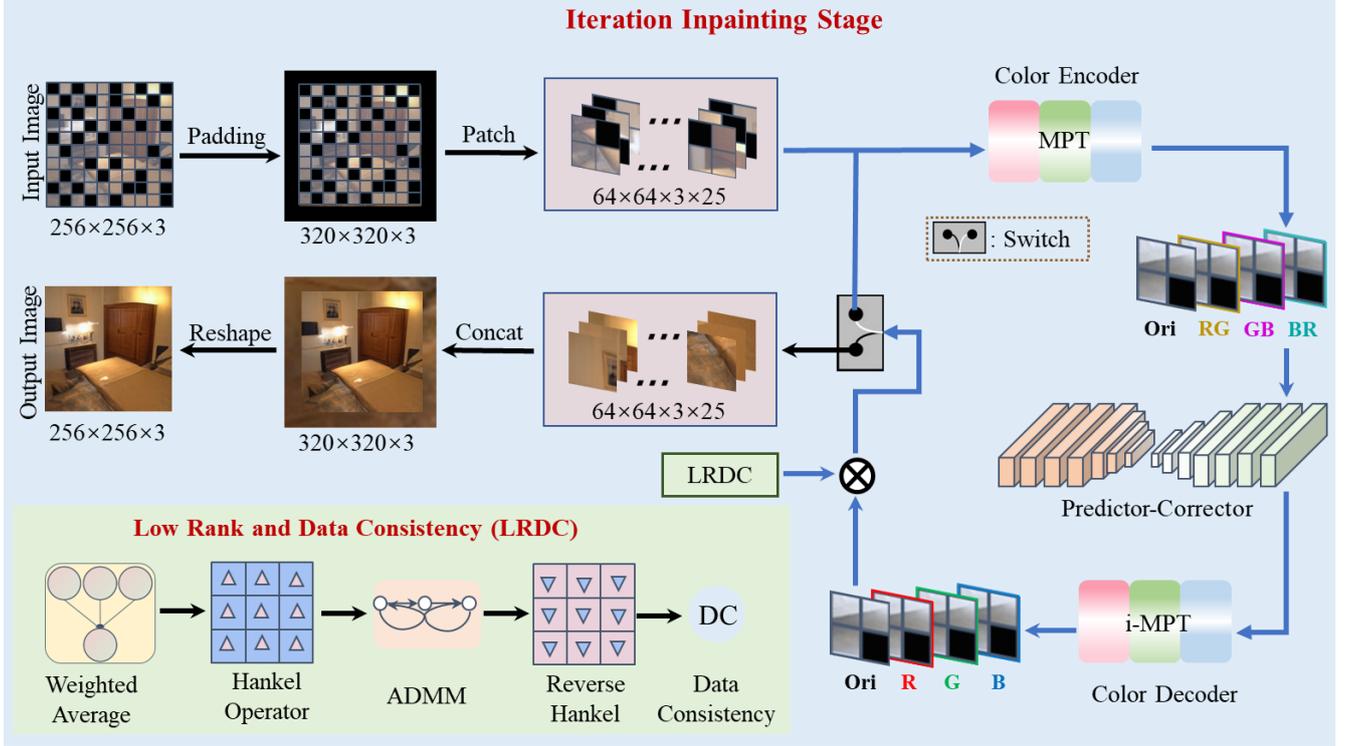

**Fig. 6.** The pipeline of iteration inpainting procedure in ESDiff. Three steps are contained in the procedure: Firstly, patches are extracted patch-by-patch from the observation after padding. Then, perform *R-G* perturbation, *G-B* perturbation, and *B-R* perturbation to obtain three different types of perturbation patches with a size of $64\times64\times3$. Finally, the PC sampler, Inverse Perturbation, Low-rank step and DC steps are performed alternatively.

For the varying color textures of each reconstructed image, the redundancy between channels differs. As previously mentioned, perturbation leverages this redundancy, enabling the image to learn additional information. However, since the channel redundancy varies for each image, we employ an adaptive inverse transform to accomplish the reconstruction task for each one.

During the experiment, we observed that the changes in weights among the three different perturbation methods, when averaged and summed to reconstruct an image significantly affect the experimental outcomes. Specifically, when the correlation coefficient between channels is high, increasing the weight between them tends to yield more visually pleasant results. However, it is important to note that perturbation between channels with low correlation coefficients is not without merit. The primary colors of different regions within a color image can vary, resulting in differing correlation coefficients across regions. Consequently, it is necessary to include a weighted summation for perturbation methods corresponding to two channels with low correlation. The SoftMax function is employed to constrain the weight range of each transformation to (0, 1), ensuring that the sum of all weights equals 1.

*Low-rank (LR) Step.* To make the low-rank step more fitting, Hankel operations are usually performed on patches before iteration. After that, these large-scale structured Hankel matrices of size of $3249\times192$ are converted into tensors with appropriate sizes for ADMM.

According to the low-rank property in structural-Hankel matrix, the formulation of low-rank constraint term in Eq. (13) is described as:

$$\min_{\mathcal{X}_{LR}} rank(\mathcal{X}_{LR}) \quad (16)$$
$$s.t. \ \mathcal{X}_{LR} = H(\mathcal{X}_{MP})$$

Subsequently, ADMM [40] is elaborated to handle Eq. (16), which can be obtain by Eq. (17)-(20):

$$U = \mu(T^{-1}(\mathcal{X}_n)+\Lambda)V(I+\mu V^H V)^{-1} \quad (17)$$
$$V = \mu(T^{-1}(\mathcal{X}_n)+\Lambda)^H U(I+\mu U^H U)^{-1} \quad (18)$$
$$\Lambda = T^{-1}(\mathcal{X}_n)-UV^H+\Lambda \quad (19)$$
$$\mathcal{X}_{LR} = U_n V^H - \Lambda \quad (20)$$

where $\mu$ denotes the step size parameter that regulates the rate of convergence, $H$ represents the transpose operator of a Hankel matrix, $T^{-1}$ is the unfolding operator, $\mathcal{X}_{LR}$ is updated using ADMM. To reduce the computational complexity, we initialize $U, V, \Lambda$ via a SVD-free algorithm proposed by LMaFit [41].

*Data Consistency (DC) Step.* After MPT, PC sampler, i-MPT, and low-rank steps, a DC step is employed to process the intermediate data. In particular, after updating the virtual variable $\mathcal{X}_{MP}$ via low-rank step, $\mathcal{X}_{LR}$ is turned back to the image patch form via the reverse-hankel operator $H^\dagger$. Plugging it into the DC formulation from Eq. (2), it yields:

$$x_n^i = \arg\min_{x_n}\left[\|Dx_n - y_n\|^2 + \lambda\|x_n - H^\dagger(\mathcal{X}_{LR})\|^2\right] \quad (21)$$

The least-square minimization in Eq. (21) can be solved

as follows:

$$x_n^i = \frac{D^T y_n + \lambda H^\dagger(\mathcal{X}_{LR})}{1+\lambda} \quad (22)$$

Here, in the noiseless setting (*i.e.*, $\lambda \to \infty$), we replace the $i^{th}$ predicted coefficient with the original coefficient.

Comprehensively, pseudo-code of ESDiff for prior learning and iterative inpainting are formally described in Alg. 1. The whole iterative inpainting procedure consists of a two-level loop: The outer loop and the inner loop perform predictor and corrector, respectively. In one iteration $i^{th}$, Inverse Perturbation, Low-rank step and DC step are incorporated into both the outer loop and the inner loops, respectively. Fig. 6 depicts the iterative inpainting process deployed by ESDiff to restore a damaged image of size $256 \times 256 \times 3$. More specifically, during the prior learning stage, patches should be extracted from the observations. These patches are subsequently perturbed using an encoding strategy with a virtual mask to enhance the generalization capability of the model. Subsequently, in the conditional generation process, the preprocessed samples are initially perturbed, followed by the updating of the preprocessed tensors through the PC sampler. Secondly, in the low-rank step, our method applies inverse-mutual perturbation transformation to the patch, processes the updated tensor using the Hankel operator, and alternately updates these matrices of size $3249 \times 192$ through the ADMM algorithm. The intermediate patches are then reconstructed from the updated Hankel matrix using the inverse Hankel operator. Finally, DC is inserted into these intermediate patches $x_n$ as input to the next iteration. When the iteration ends, the final inpainted image is obtained by concatenating all patches $\{x_n, n=1,\cdots,25\}$.

---

**Algorithm 1 ESDiff**

**Generative modeling for prior learning**

**Dataset:** Several image data $x$

1: Extracting internal-middle patches $x_n$

2: Channel perturbation on patches. $\mathcal{X}_n = MPT(x_n)$

3: Training $s_\theta(\mathcal{X},t) \simeq \nabla_{\mathcal{X}} \log p_t(\mathcal{X})$ on patches $\mathcal{X}_n$

4: Trained ESDiff

**Conditional generation for iterative inpainting**

**Initialize:** $\sigma_i, \varepsilon_i, T, N, M, U, V, \Lambda$

1: **For** $i = N-1$ to 0 **do (Outer loop)**

2: $\mathcal{X}_n = MPT(x_n)$; $\mathcal{X}_n^{\mathbb{N}} \sim \mathbb{N}(0, \sigma_T I)$ (7) **(MPT)**

3: Update $\mathcal{X}_n^i$ via Eq. (14) **(Predictor)**

4: Update $\mathcal{X}_n^i$ via $\mathcal{X}_n^i = i - MPT(\mathcal{X}_n^i)$ (8) **(i-MPT)**

5: Update $U^i, V^i, \Lambda^i, X_{LR}^i$ via Eqs. (17-20) **(LR)**

6: Update $x_n^i$ via Eq. (22) **(DC)**

7: **For** $j = 1$ to $M$ **do (Inner loop)**

8: $\mathcal{X}_n = MPT(x_n)$; $\mathcal{X}_n^{\mathbb{N}} \sim \mathbb{N}(0, \sigma_T I)$ (7) **(MPT)**

9: Update $\mathcal{X}_n^{i,j}$ via Eq. (15) **(Corrector)**

10: Update $\mathcal{X}_n^i$ via $\mathcal{X}_n^i = i - MPT(\mathcal{X}_n^i)$ (8) **(i-MPT)**

11: Update $U^{i,j}, V^{i,j}, \Lambda^{i,j}, X_{LR}^{i,j}$ via Eqs. (17-20) **(LR)**

12: Update $x_n^{i,j}$ via Eq. (22) **(DC)**

13: **End for**

14: **End for**

15: $x_{rec} = Concatenate(x_n^{i,j})$ $n = 1, 2, \cdots, 25$

16: **Return** $x_{rec}$

---

## IV. EXPERIMENTS

In this section, a set of experiments on color image restoration are presented to demonstrate the effectiveness, robustness, and flexibility of ESDiff. Particularly, the experimental implementations and datasets for evaluation are detailed.

### A. Experiment Setup

**1) *Datasets:*** All the image inpainting tests are conducted on LSUN dataset [42] and six standard natural images. Specifically, the LSUN dataset is a large-scale color image dataset with a vast number of samples and diverse image types. It comprises 10 scene categories and 20 object categories, with each category containing approximately one million sample images. For our study, we selected the indoor scene category named LSUN-bedroom, which provides a substantial number of samples (over 3 million) and a diverse range of colors, providing a robust basis for evaluating the effectiveness of ESDiff. Meanwhile, the six standard natural images—Baboon, Barbara, Boat, Cameraman, House, and Peppers—are characterized by rich color and texture, making them ideal for evaluating restoration results. It is worth noting that we employ a model trained on only 10 LSUN-bedroom images to test on both the 7 standard natural images and 100 LSUN-bedroom images. Besides, to further demonstrate the exceptional performance and robustness of ESDiff, we utilize the classic large-scale datasets, ImageNet and the Berkeley Segmentation Dataset (BSD), which encompass a diverse range of images, from natural scenes to specific objects such as plants, people, and food."

**2) *Parameter Setting:*** The pixel values of all images can be normalized to the range $[0,1]$. with the reference [28] specifically noting that the standard deviation of the noise varies over time, defined as follows:

$$\sigma(t) = \sigma_{\min}(\sigma_{\max}/\sigma_{\min})^t, \; t \in (0,1] \quad (23)$$

Based on our experience, we set $\sigma_{\min}$ to be 0.01 and $\sigma_{\max}$ to be 378. Over time, the intensity of noise gradually increases. In addition, the optimization process employs a fixed learning rate of 0.0002 with Adaptive Moment Estimation (Adam). During the iterative inpainting stage, the number of outer iterations is set to $N = 1000$. $\lambda$ is an adjustable parameter that we set to a value of 1 based on our experimental experience. Furthermore, the signal-to-noise ratio is given by $SNR = 0.075$. In each iteration of the outer loop's prediction process, the inner loop's correction step is performed once using annealed Langevin dynamics. We adopt the architecture of NCSN++, but it should be emphasized that we construct the high-dimensional representation compared to the original NCSN++ to enhance feature extraction.

**3) *Evaluation Metrics:*** In the most reported experiments, the quality of images restored by different methods is assessed using two conventional metrics: Peak Signal-to-Noise Ratio (PSNR) and Structural Similarity Index (SSIM). Generally, higher PSNR and SSIM values indicate better visual quality with more preserved details. Let $x$ Denote the reconstructed image and $\hat{x}$ to be the ground truth, the PSNR is defined as:

$$PSNR(x, \hat{x}) = 20 \log_{10}[\frac{Max(\hat{x})}{\|x - \hat{x}\|_2}] \quad (24)$$

The SSIM is defined as:

$$SSIM(x,\hat{x}) = \frac{(2\mu_x\mu_{\hat{x}} + c_1)(2\sigma_{x\hat{x}} + c_2)}{(\mu_x^2 + \mu_{\hat{x}}^2 + c_1)(\sigma_x^2 + \sigma_{\hat{x}}^2 + c_2)} \quad (25)$$

The training and experiments are performed with a customized version of Pytorch on an Intel Core i7 3.70 GHz CPU and a NVIDIA-GeForce-GTX 2080 12GB GPU. For the convenience of reproducible research, the source code of ESDiff can be downloaded from the website: https://github.com/yqx7150/ESDiff.

### B. Experimental Comparison

In this subsection, to reveal the remarkable performance and diversity of ESDiff, we test it on LSUN-bedroom dataset and six standard natural images, respectively.

**1) *Test on LSUN-bedroom Dataset*:** In this experiment, three masks are chosen as the degradation operators: Block, text, and random. To assess the effectiveness of ESDiff and test its ability to learn semantically meaningful image representations and restore damaged images to their original states, we compare its inpainting results with those obtained from kernel regression [15], K-SVD [16], ALOHA [17], NCSN++ [28], DIP [20] and ESDiff.

Fig. 7 shows a direct image comparison between ESDiff and other methods when the original image is covered by 80%. The second row shows the details of the three same locations of the restored images of all methods. The ESDiff method showed excellent image restoration performance. The images restored by other methods generally restored the outline of the original image, but the texture features were processed too smoothly and were different from the original image. Patch-based image reconstruction methods often employ averaging of overlapping patches to ensure seamless transitions between inserted patches and existing image data. This process, while effective in mitigating visual discontinuities at patch boundaries, inherently dilutes sharp edges and high-frequency details, leading to an overall smoothness in texture. The NCSN++ method made some improvements, but a certain degree of artifacts and noise were retained when processing texture and denoising amplification. For it, removing noise and recovering high-frequency texture details may be challenging, resulting in the inevitable retention of noise elements. The texture restoration effect of the ESDiff method was more satisfactory, especially in the subtle details, which were closer to the original image. ESDiff combines encoding strategies with prior learning to fully utilize the multi-scale features of images and learn complex detailed textures and image structures when training samples are in sufficient. Fig. 8 shows the case where 80% of the original image is covered. Observing the pillow part, other methods cannot fully restore the local details, and the results of boundary smoothing and noise interference will appear. By comparison, the processing results of ESDiff show more reliable results. The above results show the ability of ESDiff to effectively restore images and enhance the fidelity of local details on the LSUN-bedroom dataset.

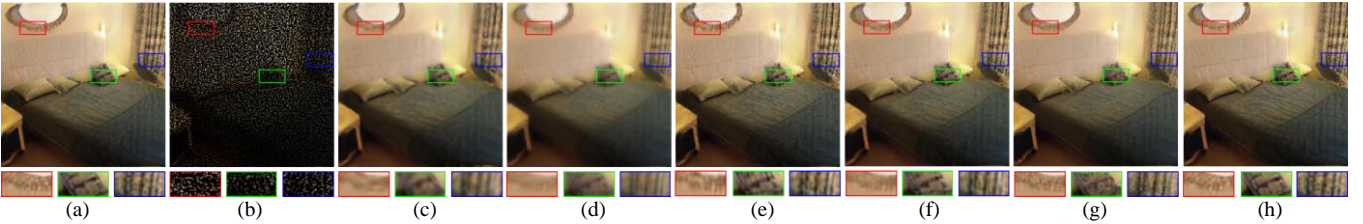

**Fig. 7.** Subjective comparison on LSUN-bedroom dataset. The results are restored from the observation under the 80% missing samples. (a) Ground truth, (b) Observation, (c) kernel regression, (d) K-SVD, (e) ALOHA, (f) DIP, (g) NCSN++, (h) ESDiff.

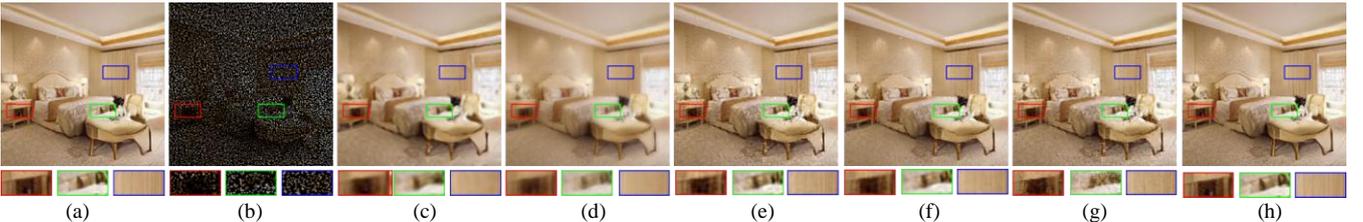

**Fig. 8.** Subjective comparison on LSUN-bedroom dataset. The results are restored from the observation, which covers block with 80%. (a) Ground truth, (b) Observation, (c) kernel regression, (d) K-SVD, (e) ALOHA, (f) DIP, (g) NCSN++, (h) ESDiff.

TABLE I
RESTORATION PSNR AND SSIM VALUES BY VARIOUS IMAGE INPAINTING ALGORITHMS FROM BLOCK, TEXT, AND RANDOM MASKS.

| Algorithm | | kernel regression [15] | K-SVD [16] | ALOHA [17] | DIP[20] | NCSN++[28] | ESDiff |
|---|---|---|---|---|---|---|---|
| Block | | 18.11/0.7191 | 27.15/0.8600 | 27.82/0.8792 | 27.90/0.8738 | 27.27/0.8556 | **28.97/0.8953** |
| Text | Text1 | 26.78/0.8435 | 29.46/0.8793 | 34.08/0.9659 | 33.95/0.9577 | 33.60/0.9569 | **36.52/0.9592** |
| | Text2 | 15.64/0.6567 | 33.09/0.9600 | 31.87/0.9710 | 36.33/0.9707 | 36.58/0.9807 | **41.74/0.9815** |
| Random | 90% | 12.01/0.5464 | 24.22/0.7160 | 24.65/0.7639 | 25.08/0.7905 | 20.65/0.5185 | **26.25/0.8008** |
| | 80% | 24.69/0.8154 | 25.70/0.7504 | 27.58/0.8540 | 27.75/0.8594 | 25.19/0.7550 | **29.08/0.8777** |
| | 70% | 26.23/0.8305 | 25.86/0.7507 | 29.56/0.8994 | 29.19/0.8846 | 27.52/0.8332 | **30.47/0.9052** |

Table I shows the quantitative comparison of ESDiff and other methods. It can be seen that ESDiff has a significant improvement in PSNR and SSIM compared with other methods. For example, when the random masking is Block, ESDiff is 1.64 dB better than the secondary best method. For the random masking test case, ESDiff is 2.03 dB, 2.81 dB, and 2.75 dB higher than the secondary best result for 90%, 80% and 70%, respectively.

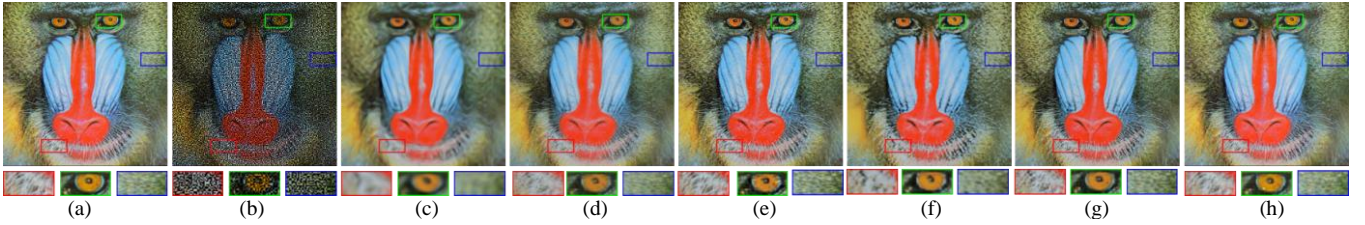

Fig. 9. Subjective comparison for restoring image Baboon. The results are restored from the observation under the 50% missing samples. (a) Ground truth, (b) Observation, (c) kernel regression, (d) K-SVD, (e) ALOHA, (f) DIP, (g) NCSN++, (h) ESDiff.

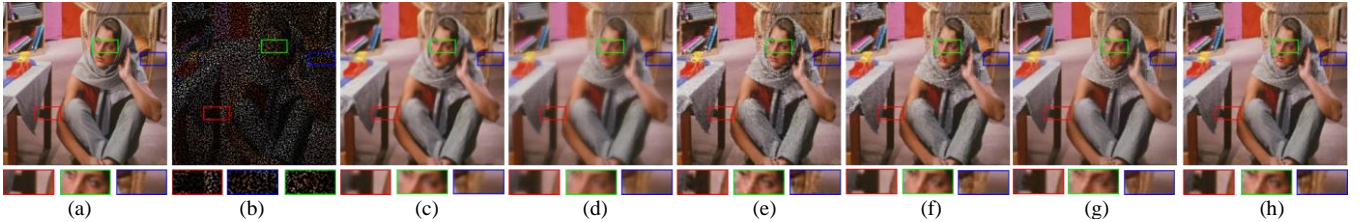

Fig. 10. Subjective comparison for recovering image Barbara. The results are restored from the observation under 80% missing samples. (a) Ground truth, (b) Observation, (c) kernel regression, (d) K-SVD, (e) ALOHA, (f) DIP, (g) NCSN++, (h) ESDiff.

**2) *Test on Six Standard Natural Images:*** To evaluate the generalization and robustness of the prior knowledge learned by ESDiff from the LUSN-bedroom dataset, we selected several standard image processing pictures, such as Baboon, Barbara, Boat, Cameraman, House and Peppers, to perform restoration experiments, where 80% (or 50%) of the pixels were randomly removed. Kernel Regression, K-SVD, ALOHA and ESDiff were selected as comparison algorithms. Figs. 9-10 show the image inpainting of Baboon with 50% masking and Barbara with 50% masking, respectively. Specifically, ESDiff does not have too much artifacts and noise in the eye area in restoration of Baboon. The K-SVD method sacrifices some texture information to reduce noise. ESDiff restores facial features of Barbara more completely and realistically, and the noise and texture are better than other methods.

In Table 2, the quantitative experimental results show that the values of some ESDiff images are not as good as those of some other methods in terms of PSNR and SSIM, but overall, ESDiff shows considerable results. When Baboon masking is 50%, ESDiff is 2.37 dB higher in PSNR and 9.40% higher in SSIM than the secondary best method K-SVD. This undoubtedly proves that image inpainting is feasible for ESDiff on 7 standard test images.

Comprehensively, the ESDiff method successfully performed image inpainting on six standard natural images. Furthermore, it is worth to highlight that the ESDiff model was trained using a dataset comprising 10 LSUN indoor bedroom images. This training set contrasts markedly with the visual scenes of the natural images used for testing. Despite these differences, the results demonstrate that ESDiff maintains excellent performance, effectively underscoring its robust capability to capture diverse features and its strong adaptability across varied visual contexts.

TABLE II
RESTORATION PSNR AND SSIM VALUES BY VARIOUS IMAGE INPAINTING ALGORITHMS FROM 80% AND 50% MISSING SAMPLES.

| Test Image | Algorithm | kernel regression[15] | K-SVD[16] | ALOHA[17] | DIP[20] | NCSN++[28] | ESDiff |
|---|---|---|---|---|---|---|---|
| Baboon | 80% | 21.90/0.5460 | 22.52/0.5694 | 20.91/0.5993 | 21.74/0.6060 | 21.85/0.5724 | **23.85/0.7005** |
|  | 50% | 22.75/0.5817 | 25.71/0.8163 | 24.75/0.8169 | 23.08/0.6848 | 24.58/0.7860 | **28.08/0.8930** |
| Barbara | 80% | 26.73/0.8151 | 26.08/0.7880 | 27.51/0.8501 | 27.18/0.8376 | 26.98/0.7992 | **29.57/0.8937** |
|  | 50% | 27.98/0.8429 | 30.19/0.9076 | 33.59/**0.9566** | 31.21/0.9231 | 31.09/0.9051 | **34.35**/0.9536 |
| Boat | 80% | 24.91/0.7475 | 25.22/0.7591 | 25.99/0.8144 | 25.45/0.8006 | 25.45/0.7620 | **26.87/0.8411** |
|  | 50% | 26.18/0.7867 | 30.06/0.9188 | 31.42/0.9397 | 31.16/0.9369 | 30.29/0.9069 | **33.43/0.9601** |
| Cameraman | 80% | 23.93/0.7965 | 23.92/0.7893 | 24.68/0.8263 | 24.18/**0.8412** | 24.40/0.7778 | **25.09**/0.8299 |
|  | 50% | 25.45/0.8246 | 28.55/0.9122 | 29.53/0.9255 | 29.50/0.9313 | 29.13/0.8889 | **30.25/0.9319** |
| House | 80% | 28.48/0.8197 | 27.67/0.8173 | 31.91/0.8723 | 31.68/0.8761 | 27.98/0.7596 | **32.65/0.8791** |
|  | 50% | 30.40/0.8429 | 32.85/0.9270 | 37.60/0.9499 | 36.90/0.9437 | 32.58/0.8635 | **38.06/0.9539** |
| Peppers | 80% | 27.75/0.8819 | 26.18/0.8419 | 28.17/0.8804 | 28.64/0.8964 | 26.08/0.7780 | **29.47/0.9035** |
|  | 50% | 29.17/0.9018 | 31.26/0.9380 | 33.79/0.9496 | 34.30/0.9561 | 31.22/0.8887 | **35.46/0.9638** |

## C. Ablation Study

In ESDiff model, MPT and LR step are key modules that will affect the efficiency of the model. In the ablation study, we firstly conduct experiments to analyze the results of different perturbations in ESDiff model and then discuss the effects of the MPT and LR modules on performance of the model though comparison experiments. As the model is trained on a small amount of datasets, we also assess the generalization ability on the proposed model on different datasets.

**1) *Different Perturbation Ratios of ESDiff:*** The generation of the virtual mask is based on a standard Gaussian distribution, with the intensity of the perturbation modulated by the perturbation ratio. Specifically, in the context of a standard Gaussian distribution, specifically when $\mu = 0$, $\sigma = 1$, a perturbation ratio of $\xi$ indicates that the probability of perturbation is given by $P(x > \xi)$. As shown in Table III, the perturbations from small to large expressed different situations, hoping to verify that a moderate perturbation ratio can optimize the model. By applying different exchange degree, the model reflects different properties. For example, when the exchange ratio increases from small to large, local irregular data is gradually obtained, and then when it decreases from large to small, global irregular data is gradually obtained. This study used a perturbation ratio of 2.81, which was the best effect, but due to the limited experimental scale, we assessed the inpainting performance on the LSUN dataset under conditions of 80% missing. In addition, the amplitude of the perturbation will also change for different data sets, and we will continue to conduct in-depth research in the future.

TABLE III
RESTORATION PSNR, SSIM AND TIME VALUES ON 80% MISSING SAMPLES BY USING DIFFERENT NUMBER OF RAW TRAINING DATA.

| Ratio | 0 | 2 | 2.81 | 3 |
|---|---|---|---|---|
| PSNR /SSIM | 28.42/0.8702 | 29.05/0.8764 | **29.08/0.8777** | 29.06/0.8770 |

**2) *Different Inpainting Strategies of Methods:*** In order to verify the ability of the ESDiff model, we conducted a series of experiments to compare the results of the ESDiff model in the reconstruction stage and the ESDiff model after a certain module was stripped off during the training stage. They are the NCSN++ model, the NCSN++ model with a low-rank module, the NCSN++ model with a perturbation module, and the ESDiff model. Table IV clearly shows that in the inpainting task under 80% masked samples, the ESDiff model has the best reconstruction effect. This is attributed to the combination of the perturbation module and the low-rank module, which is a huge improvement compared to the original NCSN++ model. The low-rank module enables the model to be more capable of inpainting the image texture, and the perturbation module strengthens the understanding of the image structure in the inpainting task. Therefore, the well-combined ESDiff model obtains the best results.

TABLE IV
RESTORATION PSNR AND SSIM VALUES ON 80% MISSING SAMPLES BY USING DIFFERENT TRAINING SCHEMES.

| Strategy | NCSN++ | NCSN++ &LR | NCSN++ &MPT | ESDiff |
|---|---|---|---|---|
| LSUN | 20.13/0.8652 | 28.42/0.8702 | 24.71/0.7057 | **29.08/0.8777** |

**3) *Different Training Datasets of ESDiff:*** Here, we train two ESDiff models using 10 images selected from the BSD and ImageNet datasets, respectively. To evaluate the performance of these models, this study supplemented the reconstruction task by applying 80% pixels masking to 100 images from the LSUN-Bedroom dataset. As reported in Table V, the restoration PSNR and SSIM values under the model learned from ImageNet are slightly better than that from BSD dataset. Accordingly, ESDiff can indeed achieve comparable results under different training datasets and merely using several raw training data.

TABLE V
RESTORATION PSNR AND SSIM VALUES BY BSD AND IMAGENET DATASETS FROM BLOCK MASKS.

| Training dataset | BSD | ImageNet |
|---|---|---|
| 100 images | 28.71/0.8735 | 28.66/0.8696 |

## V. CONCLUSION

This study proposes a diffusion model based on an encoding strategy to enhance prior learning and iterative reconstruction by perturbing the RGB channels with a virtual mask. By adding the perturbation module, the image can better learn the structure and characteristics of the image, and the data distribution of image samples can be learned with only 10 images. Furthermore, we introduced this encoding transformation into iterative inpainting and use adaptive weighted averaging to enhance image reconstruction based on the correlation between channels. Experiments have demonstrated that the ESDiff model is capable of completing good image restoration tasks with few-shot learning. Compared with other advanced methods, it has achieved good results in data rating indicators PSNR and SSIM. However, prior learning of the model necessitates paired and mutually correlated original images, which poses a significant challenge for transfer applications involving single-image channels, such as grayscale images. In the future, we will employ the powerful potential of the model to learn data distribution and apply it to other tasks to verify the strong robustness and flexibility of the model. We will continue to study the improvement of model performance by image coding strategies, continuously strengthen the diffusion model, and try it in a wider range of application scenarios.

## REFERENCES

[1] Z. Wan *et al.*, "Bringing old photos back to life," in *proceedings of the IEEE/CVF conference on computer vision and pattern recognition*, 2020, pp. 2747-2757.
[2] Z. Yi, Q. Tang, S. Azizi, D. Jang, and Z. Xu, "Contextual residual aggregation for ultra high-resolution image inpainting," in *Proceedings of the IEEE/CVF conference on computer vision and pattern recognition*, 2020, pp. 7508-7517.
[3] W. Du, H. Chen, and H. Yang, "Learning invariant representation for unsupervised image restoration," in *Proceedings of the ieee/cvf conference on computer vision and pattern recognition*, 2020, pp. 14483-14492.
[4] M. Isogawa, D. Mikami, D. Iwai, H. Kimata, and K. Sato, "Mask optimization for image inpainting," *IEEE Access,* vol. 6, pp. 69728-69741, 2018.
[5] S. M. Muddala, R. Olsson, and M. Sjöström, "Spatio-temporal consistent depth-image-based rendering using layered depth image and inpainting," *EURASIP Journal on Image and Video Processing,* vol. 2016, pp. 1-19, 2016.
[6] X. Ning, W. Li, and W. Liu, "A fast single image haze removal method based on human retina property," *IEICE TRANSACTIONS*

*on Information and Systems,* vol. 100, no. 1, pp. 211-214, 2017.
[7] Zomet, "Learning how to inpaint from global image statistics," in *Proceedings Ninth IEEE international conference on computer vision,* 2003, pp. 305-312 vol. 1: IEEE.
[8] S. Masnou and J.-M. Morel, "Level lines based disocclusion," in *Proceedings 1998 International Conference on Image Processing. ICIP98 (Cat. No. 98CB36269),* 1998, pp. 259-263: IEEE.
[9] P. Isola, J.-Y. Zhu, T. Zhou, and A. A. Efros, "Image-to-image translation with conditional adversarial networks," in *Proceedings of the IEEE conference on computer vision and pattern recognition,* 2017, pp. 1125-1134.
[10] W. Quan, R. Zhang, Y. Zhang, Z. Li, J. Wang, and D.-M. Yan, "Image inpainting with local and global refinement," *IEEE Transactions on Image Processing,* vol. 31, pp. 2405-2420, 2022.
[11] D. Zhang, Z. Liang, G. Yang, Q. Li, L. Li, and X. Sun, "A robust forgery detection algorithm for object removal by exemplar-based image inpainting," *Multimedia Tools and Applications,* vol. 77, pp. 11823-11842, 2018.
[12] M. Bertalmio, G. Sapiro, V. Caselles, and C. Ballester, "Image inpainting," in *Proceedings of the 27th annual conference on Computer graphics and interactive techniques,* 2000, pp. 417-424.
[13] Y. Wei and S. Liu, "Domain-based structure-aware image inpainting," *Signal, Image and Video Processing,* vol. 10, pp. 911-919, 2016.
[14] T. Ružić and A. Pižurica, "Context-aware patch-based image inpainting using Markov random field modeling," *IEEE transactions on image processing,* vol. 24, no. 1, pp. 444-456, 2014.
[15] H. Takeda, S. Farsiu, and P. Milanfar, "Kernel regression for image processing and reconstruction," *IEEE Transactions on image processing,* vol. 16, no. 2, pp. 349-366, 2007.
[16] M. Aharon, M. Elad, and A. Bruckstein, "K-SVD: An algorithm for designing overcomplete dictionaries for sparse representation," *IEEE Transactions on signal processing,* vol. 54, no. 11, pp. 4311-4322, 2006.
[17] K. H. Jin and J. C. Ye, "Annihilating filter-based low-rank Hankel matrix approach for image inpainting," *IEEE Transactions on Image Processing,* vol. 24, no. 11, pp. 3498-3511, 2015.
[18] F. Yao, "Damaged region filling by improved criminisi image inpainting algorithm for thangka," *Cluster Computing,* vol. 22, pp. 13683-13691, 2019.
[19] R. Rahim and S. Nadeem, "End-to-end trained CNN encoder-decoder networks for image steganography," in *Proceedings of the European conference on computer vision (ECCV) workshops,* 2018, pp. 1-6.
[20] D. Ulyanov, A. Vedaldi, and V. Lempitsky, "Deep image prior," in *Proceedings of the IEEE conference on computer vision and pattern recognition,* 2018, pp. 9446-9454.
[21] J. Ho, A. Jain, and P. Abbeel, "Denoising diffusion probabilistic models," *Advances in neural information processing systems,* vol. 33, pp. 6840-6851, 2020.
[22] J. Song, C. Meng, and S. Ermon, "Denoising diffusion implicit models," *arXiv preprint arXiv:2010.02502,* 2020.
[23] R. Rombach, A. Blattmann, D. Lorenz, P. Esser, and B. Ommer, "High-resolution image synthesis with latent diffusion models," in *Proceedings of the IEEE/CVF conference on computer vision and pattern recognition,* 2022, pp. 10684-10695.
[24] D. Pathak, P. Krahenbuhl, J. Donahue, T. Darrell, and A. A. Efros, "Context Encoders: Feature Learning by Inpainting," in *2016 IEEE Conference on Computer Vision and Pattern Recognition (CVPR),* 2016.
[25] P. Dhariwal and A. Nichol, "Diffusion models beat gans on image synthesis," *Advances in neural information processing systems,* vol. 34, pp. 8780-8794, 2021.
[26] Y. Song and S. Ermon, "Generative modeling by estimating gradients of the data distribution," *Advances in neural information processing systems,* vol. 32, 2019.
[27] Y. Song and S. Ermon, "Improved techniques for training score-based generative models," *Advances in neural information processing systems,* vol. 33, pp. 12438-12448, 2020.
[28] Y. Song, J. Sohl-Dickstein, D. P. Kingma, A. Kumar, S. Ermon, and B. Poole, "Score-based generative modeling through stochastic differential equations," *arXiv preprint arXiv:2011.13456,* 2020.
[29] A. Lugmayr, M. Danelljan, A. Romero, F. Yu, R. Timofte, and L. Van Gool, "Repaint: Inpainting using denoising diffusion probabilistic models," in *Proceedings of the IEEE/CVF conference on computer vision and pattern recognition,* 2022, pp. 11461-11471.
[30] T. Karras, M. Aittala, T. Aila, and S. Laine, "Elucidating the design space of diffusion-based generative models," *Advances in neural information processing systems,* vol. 35, pp. 26565-26577, 2022.
[31] D. L. Donoho, "Compressed sensing," *IEEE Transactions on information theory,* vol. 52, no. 4, pp. 1289-1306, 2006.
[32] L. Mancera and J. Portilla, "L0-norm-based sparse representation through alternate projections," in *2006 International Conference on Image Processing,* 2006, pp. 2089-2092: IEEE.
[33] E. J. Candes, M. B. Wakin, and S. P. Boyd, "Enhancing sparsity by reweighted ℓ 1 minimization," *Journal of Fourier analysis and applications,* vol. 14, pp. 877-905, 2008.
[34] J. Zhang, Z. Wei, and L. Xiao, "Fractional-order iterative regularization method for total variation based image denoising," *Journal of Electronic Imaging,* vol. 21, no. 4, pp. 043005-043005, 2012.
[35] P. E. Kloeden, E. Platen, P. E. Kloeden, and E. Platen, *Stochastic differential equations*. Springer, 1992.
[36] H. Wu, J. Zhou, and Y. Li, "Deep generative model for image inpainting with local binary pattern learning and spatial attention," *IEEE Transactions on Multimedia,* vol. 24, pp. 4016-4027, 2021.
[37] S. Xu, D. Liu, and Z. Xiong, "E2I: Generative inpainting from edge to image," *IEEE Transactions on Circuits and Systems for Video Technology,* vol. 31, no. 4, pp. 1308-1322, 2020.
[38] D. He, Z. Cai, D. Zhou, and Z. Chen, "Inter-Channel Correlation Modeling and Improved Skewed Histogram Shifting for Reversible Data Hiding in Color Images," *Mathematics,* vol. 12, no. 9, p. 1283, 2024.
[39] A. Block, Y. Mroueh, and A. Rakhlin, "Generative modeling with denoising auto-encoders and Langevin sampling," *arXiv preprint arXiv:2002.00107,* 2020.
[40] S. Boyd, N. Parikh, E. Chu, B. Peleato, and J. Eckstein, "Distributed optimization and statistical learning via the alternating direction method of multipliers," *Foundations and Trends® in Machine learning,* vol. 3, no. 1, pp. 1-122, 2011.
[41] Z. Wen, W. Yin, and Y. Zhang, "Solving a low-rank factorization model for matrix completion by a nonlinear successive over-relaxation algorithm," *Mathematical Programming Computation,* vol. 4, no. 4, pp. 333-361, 2012.
[42] F. Yu, A. Seff, Y. Zhang, S. Song, T. Funkhouser, and J. Xiao, "Lsun: Construction of a large-scale image dataset using deep learning with humans in the loop," *arXiv preprint arXiv:1506.03365,* 2015.